\documentclass[10pt]{article}
\usepackage[preprint]{neurips_2025}
\usepackage{natbib}
\usepackage{hyperref}
\usepackage{graphicx}
\usepackage[T1]{fontenc}
\usepackage[utf8]{inputenc}
\usepackage{comment} 

\title{Are Humans Evolved Instruction Followers? \\ An Underlying Inductive Bias Enables \\ Rapid Instructed Task Learning}

\author{%
  Anjishnu Kumar \\ 
  Amazon Alexa AI  \\ Seattle, USA  \\ 
  \texttt{anjikum@amazon.com} \\
}

\date{}

\begin{document}

\maketitle

\begin{abstract}
Human adults can often perform a novel task correctly on the first attempt after only receiving verbal or written instructions. This rapid instructed task learning (RITL) is a hallmark of human cognitive flexibility, yet its mechanisms and parallels in artificial systems remain under-explored across disciplines. In this position paper, we argue that humans possess an \emph{evolved instruction-following bias}--an inductive bias shaped by evolution to interpret and execute linguistic instructions--which critically enables fast generalization of behavior from language. This bias functions analogously to the way large language models (LLMs) leverage instruction tuning to achieve zero-shot task performance. We synthesize evidence from cognitive science, neuroscience, and machine learning research to support this hypothesis. While instruction-following in AI is currently achieved via specialized training protocols, we posit that in humans it arises as an innate cognitive architecture feature. We outline testable predictions and call for more interdisciplinary research to investigate Instruction-Following as a unifying mechanism enabling rapid task learning in both natural and artificial neural networks.
\end{abstract}

\section{Introduction}
Humans uniquely excel at learning new tasks from instructions. With minimal or no direct experience, a person can often understand a set of written or spoken directions and immediately carry out a novel task correctly \citep{cole2013ritl}. For example, an adult might successfully operate an unfamiliar software application or play a new board game after reading a manual just once. In laboratory settings, cognitive scientists term this ability \textbf{Rapid Instructed Task Learning (RITL)}-- "The ability to rapidly reconfigure our minds to perform new tasks based on instruction". RITL allows first-trial execution of novel rule sets without iterative practice, a capacity that is highly advantageous in daily life and distinguishes human learning from standard reinforcement learning (which typically requires repeated trial-and-error) \citep{sutton2018reinforcement}. Indeed, non-human animals generally cannot do the same: they require numerous reinforced trials to acquire a new behavioral task, whereas humans often succeed by instruction on the first attempt \citep{asaad1998neuron}. 

\paragraph{Cross-Disciplinary Gap.} Research on instruction following is siloed. Cognitive and developmental scientists have studied how humans learn from pedagogy and language, while neuroscientists probe the brain networks enabling flexible task sets. In parallel, AI researchers have developed instruction-tuned language models that exhibit striking abilities to perform new tasks described in natural language \citep{ouyang2022instructgpt}. However, these communities use different terminologies and frameworks, and a unifying perspective is lacking. Cognitive neuroscience studies the topic mainly tangentially, whereas AI views instruction-following largely as merely a useful model training step. We identify a critical gap in existing theoretical work: is there an underlying principle of instruction-based generalization common to both human brains and artificial models? And if so, have humans effectively been ``instruction-tuned'' by evolution?

\paragraph{Central Hypothesis.} We propose that humans possess an evolved instruction-following bias that enables rapid generalization from linguistic instructions, analogous to how instruction tuning endows LLMs with broad task flexibility. In other words, the human mind may come prepared with inductive biases and neural machinery specialized for processing inputs as instructions in a way that is different to other species. We argue that this instruction-following bias in humans is likely an evolutionary adaptation--forged by the advantages of cultural transmission and social learning--and is a key enabler of humans’ unparalleled capacity for fast instructed learning \citep{csibra2011natped,rendell2010science}. Understanding this bias can inform the design of AI systems, and conversely, insights from instruction-tuned models can shed light on cognitive and neural mechanisms of RITL. We organize the paper as follows: (i) review relevant work from cognitive science, neuroscience, and machine learning on instruction following and rapid task generalization; (ii) articulate the proposed analogy between human and AI instruction-following, highlighting evolution in humans versus explicit training in machines; (iii) compare and contrast with animal studies and (iv) discuss testable implications for both neuroscience and AI.

\section{Background and Related Work}
\subsection{Human Cognitive Capacity for Instruction Following}
 Work on human development suggests that infants are born students, equipped with biases to learn from communicative cues i.e. "natural pedagogy". According to this theory, infants innately recognize signals such as infant-directed speech as indicators that they are being taught, and they preferentially process information presented in this pedagogical context, enabling fast social learning of cognitively opaque cultural knowledge. Human children appear predisposed to follow instructions and demonstrations, allowing rapid uptake of  knowledge that ensures cultural transmission. In adults, the pinnacle of instruction-following ability is captured by RITL: learning a novel task rule or procedure from an instruction and executing it correctly on the first try \citep{cole2013ritl}. Everyday examples include learning to use a new software after reading a guide, or mentally running through steps of a new recipe from a list. Experimental psychology demonstrates that humans can achieve high first-trial performance on arbitrary tasks when those tasks are described in advance by instructions--for example, implementing arbitrary mappings between novel stimuli and responses immediately and with high accuracy \citep{meiran2012ibr,wolfensteller2012ibl}. Such behavior would be impossible without instructions--imagine learning the rules of chess by pure trial-and-error versus having them explained verbally. RITL dramatically reduces the search space of learning by directly imposing a task set via language.

Comparative evidence underscores that RITL is a rare trait. Some primates can follow extremely simple instructions. Macaques, for example, often require many practice trials to learn even simple stimulus-response associations \citep{asaad1998neuron}. In contrast, human participants facing similar novel mapping tasks can be highly accurate on the very first trial \citep{cole2013ritl}. Our closest ape relatives show a bit more prowess--the bonobo Kanzi could execute dozens of simple English commands (e.g., ``Put your ball on the pine needles'') with substantial accuracy \citep{savagerumbaugh1993monograph}. But even Kanzi’s performance is roughly equivalent to that of a human toddler. Classic experiments by \cite{horner2005causal} showed that human children faithfully reproduced arbitrary actions they saw an adult perform while completing a simple task, in contrast, chimpanzees in the same setup skipped the irrelevant steps and went straight for the goal-efficient actions. This difference is profound: it suggests humans are intrinsically wired to follow instructions/observations from others with high fidelity, whereas apes emulate the goal but filter out what seems unnecessary. Follow-up studies have found overimitation to be robust across cultures, with verbal instructions enhancing the effect. \citep{papa2021effects}. The benefits of such a capacity are evident: being able to share new skills or behaviors through communication confers adaptive value. Computational tournaments of social learning strategies show how instruction-like social learning enhances group success and the spread of knowledge \citep{rendell2010science}. This likely created selective pressure for an instruction-following bias in the human lineage \citep{csibra2011natped}.

\paragraph{Dogs as heterospecific instruction followers.}
Domestic dogs provide a natural experiment on whether an instruction-following bias (IF-bias) can be shaped by selection. In imperative object–choice tasks, dogs reliably use human pointing and gaze to select the requested item, whereas chimpanzees often perform at chance without extensive training \citep{Hare2002,Kirchhofer2012}. Dogs pups are able to outperform wolf pups of similar age on similar tasks \citep{viranyi2008comprehension}. Taken together, the canids illustrate that selective pressure (human domestication and social hierarchy) can create a robust IF-bias for human signals — without the broad, compositional, language-mediated task recomposition that characterizes human RITL. 

\subsection{Neural Mechanisms: Task-Set Reconfiguration and Cognitive Control}
What happens in the brain when we convert an instruction into action? Cognitive neuroscience points to the frontoparietal cognitive control network--especially lateral prefrontal cortex (LPFC)--as key for implementing novel task sets on the fly \citep{duncan2010md}. Neuroimaging of RITL finds that when people encode new instructions, dorsolateral/anterior PFC represents the intended task even before any practice, and then interacts with parietal and task-relevant areas to guide execution \citep{cole2011rapid,cole2013flexhub}. This aligns with task-set reconfiguration: when faced with a new instructed task, the brain rapidly configures a network that implements the mapping from stimuli to responses as specified by the rule; the pattern of frontoparietal connectivity shifts flexibly with current demands and can predict which task a person is performing, consistently across novel and practiced tasks \citep{cole2013flexhub}. These ``flexible hubs'' are thought to underwrite generalized cognitive control.

There are indications that humans’ proficiency at instruction-based task-set formation has distinct neural underpinnings compared to other species. Anatomically, anterior prefrontal cortex (approx.\ Brodmann area 10) is disproportionately enlarged in humans relative to other primates, and among non-human apes the bonobo shows a relatively large area 10 \citep{semendeferi2001area10}. This suggests evolutionary expansion of prefrontal circuitry supporting abstract, instruction-derived rules. Consistently, classic theory casts PFC as an instruction parser/buffer transforming linguistic/symbolic instruction into an executable task set via top-down control \citep{miller2001pfc}. Overall, the evidence suggests a general-purpose task engine configurable by instruction.

\subsection{Instruction Tuning in AI and Zero-Shot Task Generalization}
Mirroring the human ability to generalize, modern large language models have recently demonstrated remarkable zero-shot and few-shot capabilities. Base pretrained models (e.g., GPT-3) can solve new tasks given a few examples or prompts, but unable to follow arbitrary user instructions out-of-the-box \citep{brown2020gpt3}. 

To bridge this, instruction tuning fine-tunes models on many natural-language (instruction, response) pairs spanning diverse tasks (translation, QA, arithmetic, etc.), explicitly teaching models to follow instructions. Through supervised learning (often with alignment via human feedback), parameters are adjusted so the model interprets the instruction and generates an appropriate completion \citep{ouyang2022instructgpt}. The outcome is strong zero-shot performance on unseen tasks--sometimes matching or exceeding larger non-tuned models or few-shot baselines simply by understanding the instruction \citep{wei2022flan,chung2022scaling}. In effect, instruction tuning confers task flexibility akin to human RITL. These developments motivate viewing instruction tuning as a machine analog of human instruction following.

\section{Evolved Instruction-Following Bias: Linking Human and Machine}
We hypothesize that the human brain comes equipped with a specialized inductive bias for interpreting and applying instructions, selected through the evolution of communication and culture. By default, when a human hears or reads a directive (especially in a pedagogical context), the cognitive system attempts to integrate that information as a guide for behavior--an innate expectation that language conveys actionable, generalized knowledge \citep{csibra2011natped}. This is analogous to how an instruction-tuned LLM exhibits an internal policy (learned via fine-tuning) to treat inputs as instructions and produce appropriate outputs \citep{sanh2022t0,ouyang2022instructgpt}. For the LLM, the skill is externally added by additional training; for humans, we propose it is deeply ingrained, a result of our neurological ``hardware'' and inherited cognitive architecture. From an evolutionary perspective, individuals who could rapidly share survival skills or tool-use techniques via language would have an advantage \citep{rendell2010science}. Over generations, this could favor brains adept at rapidly mapping linguistic input to goal-directed actions. Natural pedagogy in infants and the jump in instruction-following capacities from monkeys to apes to humans both point to evolutionary tuning toward instruction-based learning. In short, what instruction tuning is to GPT-3, evolution (and cultural evolution) is to \emph{Homo sapiens}: a process endowing the system with a general skill of following instructions.

This perspective reframes instruction following as a core computational strategy underlying human cognitive flexibility. It suggests that the brain’s default prior (especially in cooperative social contexts) is that new behaviors can be acquired by listening and understanding, not solely by direct experience. Just as an instruction-tuned model combines pretraining knowledge with the prompt to produce the solution, a human combines prior world knowledge with the content of instructions to synthesize a plan for a novel task. The hypothesis further implies that certain neural mechanisms may be dedicated to, or disproportionately used for, parsing and implementing instructions (e.g., specialized representations of instructions in PFC) \citep{miller2001pfc,cole2011rapid}. Recent work by \citep{goldstein2024alignment}  shows alignment between language-model embeddings and human neural representations during language processing, and work by \citep{aw2023instruction} demonstates that instruction-tuning of LLMs improves alignment between language-model and brain representations.

\section{Implications and Future Directions}
If humans indeed have an evolved instruction-following bias analogous to AI instruction tuning, several testable predictions and research directions emerge:

\paragraph{Neural Representation of Instructions.} The hypothesis predicts that the brain maintains explicit representations of intended task instructions, likely in frontal cortex, detectable during encoding and predictive of later performance. During the instruction-encoding phase (before execution), one should observe stable LPFC patterns corresponding to the rule, which then configure broader task networks. Identifying alignments between instruction semantics and neural state geometry would support a dedicated instruction-processing mechanism \citep{goldstein2024alignment}.

\paragraph{Flexible-Hub Dynamics.} If instruction following relies on flexible hubs in frontoparietal networks, perturbations should selectively impair instruction-based setup while sparing well-learned tasks. For instance, transient disruption of LPFC during instruction presentation should reduce the ability to configure new task sets, paralleling classic evidence that LPFC disruption impairs top-down control \citep{miller2001pfc}. 

\paragraph{Comparative and Developmental Studies.} The bias should be present early in development and rudimentary or absent in non-human animals. Experiments can test toddlers’ one-shot instruction following in novel tasks, tracking growth in parallel with language ability, and compare with primates using simplified gestural/symbolic instructions. 

\paragraph{Parallels in AI Behavior.} Instruction-tuned models offer a tool to simulate and predict human instruction following. By presenting matched novel tasks to humans and tuned LLMs; comparing divergences can reveal where human inductive priors differ from current models.

\section{Conclusion}
We argue that RITL should be regarded as a foundational aspect of human intelligence, fostered by evolution and cultural transmission. This instruction-following bias in humans plays a role analogous to instruction tuning in artificial language models. Recognizing this analogy provides a conceptual bridge between cognitive neuroscience and AI. For AI researchers, it has been key to creating flexible, human-like learners. For neuroscientists, it highlights instruction-following as a distinct cognitive function worthy of further focused investigation, with its own neural signatures and evolutionary history.  Interdisciplinary collaboration leading better understanding instruction-following may prove key to a unifying theory for rapid generalization.

\bibliographystyle{plainnat}
\bibliography{references}

@article{cole2013ritl,
  title={Rapid instructed task learning: A new window into the human brain's unique capacity for cognitive control},
  author={Cole, Michael W. and Laurent, Patryk and Stocco, Andrea},
  journal={Frontiers in Human Neuroscience},
  volume={7},
  pages={207},
  year={2013},
  doi={10.3389/fnhum.2013.00207},
  url={https://pmc.ncbi.nlm.nih.gov/articles/PMC3557598/}
}

@article{duncan2010md,
  title={The multiple-demand (MD) system of the primate brain: mental programs for intelligent behaviour},
  author={Duncan, John},
  journal={Trends in Cognitive Sciences},
  volume={14},
  number={4},
  pages={172--179},
  year={2010},
  doi={10.1016/j.tics.2010.01.004}
}

@article{cole2013flexhub,
  title={Multi-task connectivity reveals flexible hubs for adaptive task control},
  author={Cole, Michael W. and Reynolds, John R. and Power, Jonathan D. and Repov{\v{s}}, Grega and Anticevic, Alan and Braver, Todd S.},
  journal={Proceedings of the National Academy of Sciences},
  volume={110},
  number={28},
  pages={10061--10066},
  year={2013},
  doi={10.1073/pnas.1302916110}
}

@article{cole2011rapid,
  title={Rapid transfer of abstract rules to novel contexts in human lateral prefrontal cortex},
  author={Cole, Michael W. and Etzel, Joset A. and Zacks, Jessica M. and Schneider, Walter and Braver, Todd S.},
  journal={Frontiers in Human Neuroscience},
  volume={5},
  pages={142},
  year={2011},
  doi={10.3389/fnhum.2011.00142},
  url={https://www.frontiersin.org/articles/10.3389/fnhum.2011.00142/full}
}

@article{miller2001pfc,
  title={An integrative theory of prefrontal cortex function},
  author={Miller, Earl K. and Cohen, Jonathan D.},
  journal={Annual Review of Neuroscience},
  volume={24},
  pages={167--202},
  year={2001},
  doi={10.1146/annurev.neuro.24.1.167}
}

@article{csibra2011natped,
  title={Natural pedagogy as evolutionary adaptation},
  author={Csibra, Gergely and Gergely, Gy{\"o}rgy},
  journal={Philosophical Transactions of the Royal Society B},
  volume={366},
  number={1567},
  pages={1149--1157},
  year={2011},
  doi={10.1098/rstb.2010.0319},
  url={https://pmc.ncbi.nlm.nih.gov/articles/PMC3049090/}
}

@article{wolfensteller2012ibl,
  title={Frontostriatal mechanisms in instruction-based learning},
  author={Wolfensteller, Uta and Ruge, Hannes},
  journal={Frontiers in Psychology},
  volume={3},
  pages={192},
  year={2012},
  doi={10.3389/fpsyg.2012.00192},
  url={https://www.frontiersin.org/articles/10.3389/fpsyg.2012.00192/full}
}

@article{meiran2012ibr,
  title={When planning results in loss of control: intention-based reflexivity and working memory},
  author={Meiran, Nachshon and Pereg, Maayan and Kessler, Yoav and Cole, Michael W. and Braver, Todd S.},
  journal={Frontiers in Human Neuroscience},
  volume={6},
  pages={295},
  year={2012},
  doi={10.3389/fnhum.2012.00295}
}

@article{asaad1998neuron,
  title={Neural activity in the primate prefrontal cortex during associative learning},
  author={Asaad, Wael F. and Rainer, Gregor and Miller, Earl K.},
  journal={Neuron},
  volume={21},
  number={6},
  pages={1399--1407},
  year={1998},
  doi={10.1016/S0896-6273(00)80658-3}
}

@book{savagerumbaugh1993monograph,
  title={Language Comprehension in Ape and Child},
  author={Savage-Rumbaugh, E. Sue and Murphy, Jeannine and Sevcik, Rose A. and Brakke, Karen E. and Williams, Shelly L. and Rumbaugh, Duane M.},
  series={Monographs of the Society for Research in Child Development},
  volume={58},
  number={3/4},
  year={1993},
  publisher={Wiley},
  url={https://www.bonobohope.org/pdf/1993-Language-Comprehension-Ape.pdf}
}

@article{semendeferi2001area10,
  title={Prefrontal cortex in humans and apes: A comparative study of area 10},
  author={Semendeferi, Katerina and Armstrong, Estel and Schleicher, Axel and Zilles, Karl and Van Hoesen, Gary W.},
  journal={American Journal of Physical Anthropology},
  volume={114},
  number={3},
  pages={224--241},
  year={2001},
  doi={10.1002/1096-8644(200103)114:3<224::AID-AJPA1022>3.0.CO;2-I},
  url={https://cs.brown.edu/people/tdean/projects/cortex/course/suggested_reading_list/supplements/documents/SemendeferietalAJPA-01.pdf}
}

@article{rendell2010science,
  title={Why copy others? Insights from the social learning strategies tournament},
  author={Rendell, Luke and Boyd, Robert and Cownden, Daniel and Enquist, Magnus and Eriksson, Kimmo and Feldman, Marcus W. and Fogarty, Laurel and Ghirlanda, Stefano and Lillicrap, Timothy and Laland, Kevin N.},
  journal={Science},
  volume={328},
  number={5975},
  pages={208--213},
  year={2010},
  doi={10.1126/science.1184719},
  url={https://pmc.ncbi.nlm.nih.gov/articles/PMC2989663/}
}

@article{goldstein2024alignment,
  title={Alignment of brain embeddings and artificial contextual embeddings in natural language points to common geometric patterns},
  author={Goldstein, Ariel and Grinstein-Dabush, Avigail and Schain, Mariano and Wang, Haocheng and Hong, Zhuoqiao and Aubrey, Bobbi and Nastase, Samuel A. and Zada, Zaid and Ham, Eric and Feder, Amir and Gazula, Harshvardhan and Buchnik, Eliav and others},
  journal={Nature Communications},
  volume={15},
  number={1},
  year={2024},
  doi={10.1038/s41467-024-46631-y},
  url={https://ideas.repec.org/a/nat/natcom/v15y2024i1d10.1038_s41467-024-46631-y.html}
}

@inproceedings{brown2020gpt3,
  title={Language Models are Few-Shot Learners},
  author={Brown, Tom B. and Mann, Benjamin and Ryder, Nick and Subbiah, Melanie and Kaplan, Jared and Dhariwal, Prafulla and Neelakantan, Arvind and Shyam, Pranav and Sastry, Girish and Askell, Amanda and others},
  booktitle={Advances in Neural Information Processing Systems (NeurIPS)},
  year={2020},
  url={https://arxiv.org/abs/2005.14165}
}

@inproceedings{ouyang2022instructgpt,
  title={Training language models to follow instructions with human feedback},
  author={Ouyang, Long and Wu, Jeff and Jiang, Xu and Almeida, Diogo and Wainwright, Carroll L. and Mishkin, Pamela and Zhang, Chong and Agarwal, Sandhini and Slama, Katarina and Ray, Alex and Schulman, John and Hilton, Jacob and Kelton, Fraser and Miller, Luke and Simens, Maddie and Askell, Amanda and Welinder, Peter and Christiano, Paul and Leike, Jan and Lowe, Ryan},
  booktitle={Advances in Neural Information Processing Systems (NeurIPS)},
  year={2022},
  url={https://proceedings.neurips.cc/paper_files/paper/2022/file/b1efde53be364a73914f58805a001731-Paper-Conference.pdf}
}

@inproceedings{wei2022flan,
  title={Finetuned Language Models Are Zero-Shot Learners},
  author={Wei, Jason and Bosma, Maarten and Zhao, Vincent and others},
  booktitle={International Conference on Learning Representations (ICLR)},
  year={2022},
  url={https://arxiv.org/abs/2109.01652}
}

@article{chung2022scaling,
  title={Scaling Instruction-Finetuned Language Models},
  author={Chung, Hyung Won and Hou, Le and Longpre, Shayne and Zoph, Barret and Tay, Yi and Fedus, William and Li, Yunxuan and Wang, Xuezhi and Dehghani, Mostafa and Brahma, Siddhartha and others},
  journal={arXiv preprint arXiv:2210.11416},
  year={2022},
  url={https://arxiv.org/abs/2210.11416}
}

@inproceedings{sanh2022t0,
  title={Multitask Prompted Training Enables Zero-Shot Task Generalization},
  author={Sanh, Victor and Webson, Albert and Raffel, Colin and Bach, Stephen H. and Sutawika, Lintang and Alyafeai, Zaid and Chaffin, Antoine and Stiegler, Arnaud and Le Scao, Teven and Raja, Arun and others},
  booktitle={International Conference on Learning Representations (ICLR)},
  year={2022},
  url={https://arxiv.org/abs/2110.08207}
}

@book{sutton2018reinforcement,
  title={Reinforcement Learning: An Introduction (2nd ed.)},
  author={Sutton, Richard S. and Barto, Andrew G.},
  publisher={MIT Press},
  year={2018},
  url={http://incompleteideas.net/book/the-book-2nd.html}
}

@article{horner2005causal,
  title={Causal knowledge and imitation/emulation switching in chimpanzees (Pan troglodytes) and children (Homo sapiens)},
  author={Horner, Victoria and Whiten, Andrew},
  journal={Animal cognition},
  volume={8},
  number={3},
  pages={164--181},
  year={2005},
  publisher={Springer}
}

@article{papa2021effects,
  title={Effects of verbal instruction vs. modelling on imitation and overimitation},
  author={Papa, Aliki and Cristea, Mioara and McGuigan, Nicola and Tamariz, Monica},
  journal={Humanities and Social Sciences Communications},
  volume={8},
  number={1},
  pages={1--12},
  year={2021},
  publisher={Palgrave}
}

@article{Hare2002,
  author = {Hare, B. and Brown, M. and Williamson, C. and Tomasello, M.},
  title = {The domestication of social cognition in dogs},
  journal = {Science},
  year = {2002},
  volume = {298},
  number = {5598},
  pages = {1636--1639},
  doi = {10.1126/science.1072702}
}

@article{Kirchhofer2012,
  author = {Kirchhofer, K. C. and Zimmermann, F. and Kaminski, J. and Tomasello, M.},
  title = {Dogs (Canis familiaris), but not chimpanzees (Pan troglodytes), understand imperative pointing},
  journal = {PLOS ONE},
  year = {2012},
  volume = {7},
  number = {2},
  pages = {e30913},
  doi = {10.1371/journal.pone.0030913}
}

@article{viranyi2008comprehension,
  title={Comprehension of human pointing gestures in young human-reared wolves (Canis lupus) and dogs (Canis familiaris)},
  author={Vir{\'a}nyi, Zs{\'o}fia and G{\'a}csi, M{\'a}rta and Kubinyi, Enik{\H{o}} and Top{\'a}l, J{\'o}zsef and Bel{\'e}nyi, Beatrix and Ujfalussy, Dorottya and Mikl{\'o}si, {\'A}d{\'a}m},
  journal={Animal cognition},
  volume={11},
  number={3},
  pages={373--387},
  year={2008},
  publisher={Springer}
}

@article{aw2023instruction,
  title={Instruction-tuning aligns llms to the human brain},
  author={Aw, Khai Loong and Montariol, Syrielle and AlKhamissi, Badr and Schrimpf, Martin and Bosselut, Antoine},
  journal={arXiv preprint arXiv:2312.00575},
  year={2023}
}
\end{document}